\pdfoutput=1

\documentclass[11pt]{article}

\usepackage{emnlp2021}

\usepackage{times}
\usepackage{latexsym}

\usepackage[T1]{fontenc}

\usepackage[utf8]{inputenc}

\usepackage{microtype}

\usepackage{microtype}

\usepackage{url}
\usepackage{latexsym}

\usepackage{graphicx}
\usepackage{tabularx}
\usepackage{soul}
\usepackage{varioref}
\usepackage{hyperref}
\usepackage{rotating}
\usepackage{epstopdf}
\usepackage{xstring}

\title{We Need to Talk About Data: The Importance of Data Readiness in Natural Language Processing}

\author{Fredrik Olsson\thanks{{ }{ }The lion's share of the work was carried out while at RISE, Research Institutes of Sweden.}\\
  Gavagai \\ 
  Sweden \\
  {\tt fredrik.olsson@gavagai.io} 
  \And
  Magnus Sahlgren\footnotemark[1] \\
  AI Sweden \\ 
  Sweden \\
  {\tt magnus.sahlgren@ai.se} 
  \\}

\date{}

\begin{document}
\maketitle
\begin{abstract}
In this paper, we identify the state of data as being an important reason for failure in applied Natural Language Processing (NLP) projects. 
We argue that there is a gap between academic research in NLP and its application to problems outside academia, and that this gap is rooted in poor mutual understanding between academic researchers and their non-academic peers who seek to apply research results to their operations. To foster transfer of research results from academia to non-academic settings, and the corresponding influx of requirements back to academia, we propose a method for improving the communication between researchers and external stakeholders regarding the accessibility, validity, and utility of data based on Data Readiness Levels \cite{lawrence2017data}. While still in its infancy, the method has been iterated on and applied in multiple innovation and research projects carried out with stakeholders in both the private and public sectors. Finally, we invite researchers and practitioners to share their experiences, and thus contributing to a body of work aimed at raising awareness of the importance of data readiness for NLP.
\end{abstract}

\section{Introduction}
\label{sec:intro}


NLP has always been an applied discipline, with inspiration drawn both from basic reseach in computational linguistics, computer science, and cognitive science, as well as from business problems and applications in industry and the public sector. Even if some NLP researchers prefer to work at lower Technology Readiness Levels (TRLs), while others operate at a fairly high TRL range, most of the work in NLP has the potential to climb the TRL scale up to the more practical levels (i.e. from TRL level 7 upwards) \cite{NasaTrl2017}.

For those of us who work with external clients and habitually deliver results in the form of demonstrators and prototypes at TRL 6 or 7, a consistent and significant challenge is the state of the data available. In our experience, challenges regarding data are much more common in client-facing projects than challenges relating to the technical nature of models or algorithms. We argue that the lack of readiness with respect to data has become a serious obstacle when transferring findings from research to an applied setting. Even if the research problem is sufficiently well defined, and the business value of the proposed solution is well described, it is often not clear what type of data is required, if it is available, or if it at all exists.

The border between academic research in NLP and the application of the research in practical non-academic settings is becoming increasingly blurred with the convergence of NLP research to more or less production-ready frameworks and implementations. 
On the one hand, research results have never been more accessible, and it has never been easier to obtain and adjust the architecture of, e.g., a state of the art language model to accommodate a new use case, and to construct a prototype showing the value the model would contribute to an external stakeholder. 
On the other hand, while the technical maturity of the research community has improved, the understanding of business value, and by extension, the understanding of the impact and importance of data are still largely lacking. It is our firm belief that in order for NLP, and in particular the research community that targets the lower TRLs, to become even more relevant and thus also benefit from feedback from parties outside the field, we have to assume a more holistic approach to the entire life cycle of applied research, with a particular eye on {\bf data readiness}.

The intention for this paper is therefore to raise awareness of data readiness for NLP among researchers and practitioners alike, and to initiate and nurture much-needed discussions with respect to the questions that arise when addressing real-world challenges with state of the art academic research.

\begin{figure}
    \centering
    \includegraphics[width=\columnwidth]{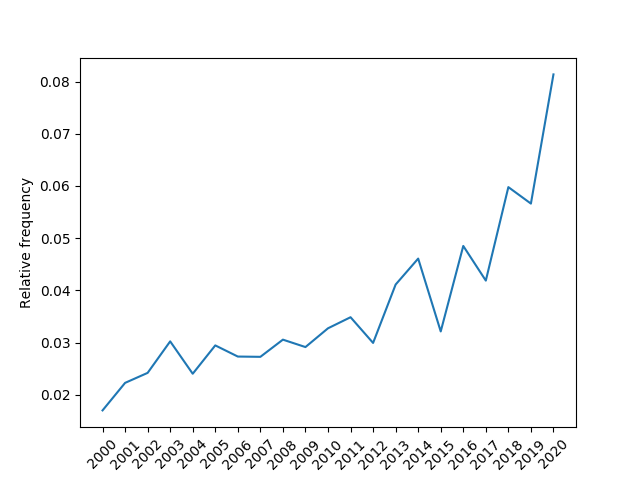}
    \caption{Relative frequency of publications in the ACL Anthology that mention the term ``data'' in the title for the last 20 years.}
    \label{fig:data}
\end{figure}

\section{Related work}
\label{sec:related-work}

As is evident from Figure \ref{fig:data}, which shows the relative frequency of publications in the ACL Anthology\footnote{\url{www.aclweb.org/anthology/}} that mention the term ``data'' in the title over the last 20 years, there is an increasing interest in questions relating to data within our field. While there are a lot of activity related to data in the research community, few attempts at fostering a discussion of the whole process, from business problem to data access, has been made. Relevant areas of academic research include the following.

{\bf Access to unlabelled training data.} Efforts to collect and distribute text data at scale include corpora originating from CommonCrawl, e.g., \cite{cc:MackenzieBenhamPetriTrippasEtAl:2020:cc-news-en,cc:ElKishkyChaudharyGuzmanKoehn:2020:ccaligned}, along with tools to facilitate corpus creation \cite{wenzek_ccnet_2020}, as well as academic initiatives such as ELRA\footnote{\url{http://www.elra.info/}} and LDC\footnote{\url{https://www.ldc.upenn.edu/}}.

{\bf Creation of labelled training data.} Research in Active Learning, e.g., \cite{settles_active_2012,siddhant_deep_2018,liang_alice_2020,ein-dor_active_2020}, as well as Zero and Few-shot Learning \cite{srivastava_zero-shot_2018,ye_zero-shot_2020,pelicon_zero-shot_2020} allows for the utilization of pre-compiled knowledge, and human-in-the-loop approaches to efficient data labelling. However, these approaches assume that there is a clear objective to address, and that unlabelled data and expertise are available. In our experience, this is rarely the case.

{\bf Bias, transparency, fairness} are all areas that have bearing towards the state of data. Much of the research, however, is concerned with situations in which the data has already been collected. Most notable recent efforts include
The {\em Dataset Nutrition Label}, which is a diagnostic framework for enabling standardized data analysis \cite{holland_dataset_2018}; {\em Data Statements for NLP} that allows for addressing exclusion and bias in the field \cite{bender_data_2018}; {\em FactSheets} intended for increasing consumers' trust in AI services \cite{arnold_factsheets_2019}; and {\em Datasheets for Datasets} for facilitating better communication between dataset creators and consumers \cite{gebru_datasheets_2020}.

{\bf Model deployment.}
Tangential to our efforts to provide stakeholders with prototypes at TRL 6-7 is the work of deploying machine learning models to a production environment. Research in this area that also touches on data readiness in some form include that of \citet{polyzotis_data_2018} and \citet{paleyes_challenges_2020}.

Work on {\bf data readiness} related to other modalities than text include \citet{van2019quality} and \citet{harvey2019standardised} that both deal with data quality in medical imaging.
\citet{austin_path_2018} outlines practical solutions to common problems with data readiness when integrating diverse datasets from heterogeneous sources. \citet{afzal_data_2020} introduces the concept {\em Data Readiness Report} as a means to document data quality across a range of standardized dimensions.

We have not found any work that focuses specifically on data readiness in the context of NLP. Our contribution is therefore a set of questions that we have found valuable to bring up in discussions with new stakeholders in order to allow us, and them, to form an understanding of the state of the data involved in the particular challenge.

\section{Data Readiness Levels}

\begin{figure*}[!ht]
\begin{center}
\includegraphics[width=0.85\linewidth]{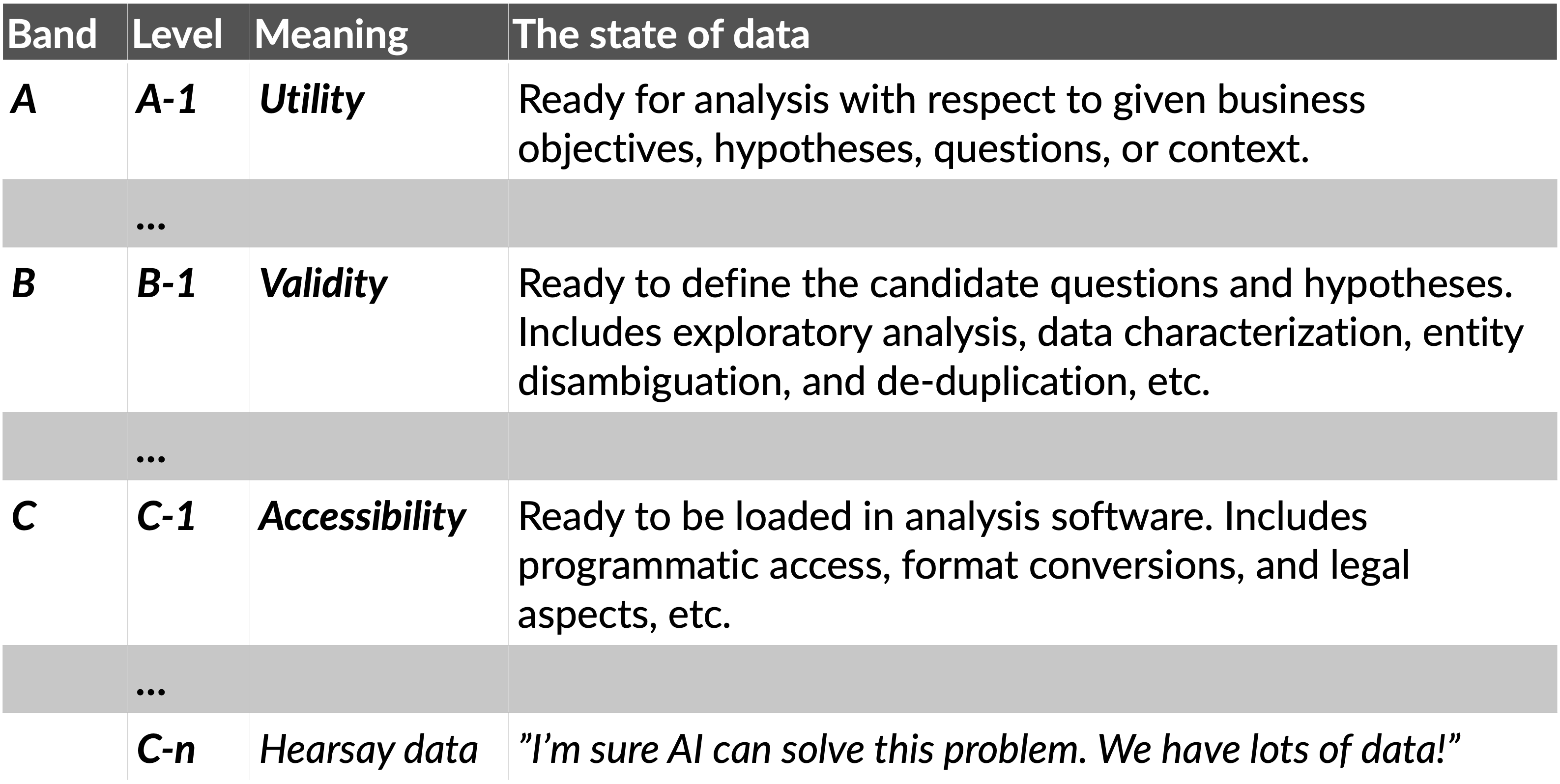}
\caption{An overview of the different bands of Data Readiness Levels.}
\label{fig:drl-table}
\end{center}
\end{figure*}

The notion of Data Readiness Levels (DRLs) provides a way of talking about data much in the same way TRLs facilitate communication regarding the maturity of technology \cite{lawrence2017data}. DRLs is a framework suitable for exchanging information with stakeholders regarding data {\em accessibility}, {\em validity}, and {\em utility}. 
There are three different major Bands of the DRLs, and each band can be thought of as consisting of multiple levels. The state of data is usually a progress from Band C towards Band A, with a particular business goal in mind. Figure \ref{fig:drl-table} illustrates the three bands of the Data Readiness Levels.

{\bf Band C} concerns the {\em accessibility} of data. All work at this level serves to grant the team intended to work with the data access to it; once access is provided, the data is considered to be at Band C - Level C-1, and ready to be brought into Band B. Issues that fall under Band C include: the existence of data; format conversion and encoding; legal aspects of accessibility; and programmatic aspects of accessibility.

{\bf Band B} concerns the {\em validity} of data. In order to pass Band B, the data has to be valid in the sense that it is representative of the task at hand. Furthermore, the data should be deduplicated, noise should be identified, missing values should be characterized, etc. At the top level of Band B, the data should be suitable for exploratory analysis, and the forming of working hypotheses.

{\bf Band A} concerns the {\em utility} of data. The utility of the data concerns the way in which the data is intended to be used: Is the data enough to solve the task at hand? A project should strive for data readiness at Band A - Level A-1. Note that the Data Readiness Levels should be interpreted with respect to a given task. 

\section{Examples of challenges}
\label{sec:examples}
The following are examples of typical challenges we have encountered, framed as belonging to the different DRLs.\footnote{The examples are deliberately kept vague since we do not want to disclose the corresponding external stakeholder.}

\subsection{DRL Band C -- Accessibility}

{\bf Example 1: Data licensing.} It was assumed that the data to work with in the project was in the public domain and readily available. It turned out the data was a proprietary news feed under license restrictions. The consequences of this were two-fold: not having access to the data generation process meant we could not address one of the stakeholder's major problems (de-duplication, relevance assessment); and, the license restrictions prevented the project from publishing the dataset along with the research findings.

{\bf Example 2: Company culture.} The ownership of data was clearly specified, but the staff did not adhere to management's request to release the data due to the uncertainty of the result of the project. This resulted in delays. The fear of the loss of jobs may impact availability of data -- data readiness depends on the overall introduction of data-driven techniques in a new organization.

{\bf Example 3: Data format.} Raw data was stored as PDF files, generated by different sources. PDF is an output format, not an input format. Projects working with PDF files will always face challenges having to do with data conversion since there is currently no way of reliably converting an arbitrary PDF file into a format useful for NLP.

\subsection{DRL Band B -- Validity}

{\bf Example 4: Annotation guidelines.} The existing annotation guidelines were elaborate, but fell short in practical applicability. Partly due to the misalignment between the annotation task and the guidelines, it became very time consuming to annotate data which resulted in a small dataset to work with. In turn, this affected the range of possible NLP techniques applicable to the problem at hand. Annotation guidelines have to be unambiguous, precise, and possible for an annotator to remember.

{\bf Example 5: Annotation quality.} The data was assumed to be of good quality, but additional investigations revealed a low inter-annotator agreement. The consequence was that the existing data could not be trusted, and the annotation work had to be re-done. If the definition of a task is too hard for human annotators to agree on, a machine trained on the data will perform poorly.

{\bf Example 6: Annotation quality.} Existing information produced by the stakeholder was assumed to be useful in creating an annotated dataset for the specific task at hand, but it turned out that the information was incomplete and insufficient. The consequence of not being able to leverage existing data for distant supervision was that the range of applicable techniques for addressing the stakeholder's problem became severely limited.

\subsection{DRL Band A -- Utility}

{\bf Example 7: Annotation expectations.} It was known that the data to work with was annotated, but the way the annotations had been made had not been communicated. Instead of sequence level annotations, the data was annotated at the document level. As a consequence, we could not explore the type of information extraction techniques we had expected, but had to resort to document classification instead.

{\bf Example 8: Data sparseness.} The overall amount and velocity of data were assumed to be of sufficient quantity, but when aligning data availability with use case requirements, it turned out the data was too sparse. The task could not be pursued.

{\bf Example 9: Project scope.} The stakeholder's team and the unannotated data they provided to the project were at an exceptionally high DRL, but annotations for training, validation, and testing were very hard to obtain since the project had not planned for annotation work. As a consequence, we implemented a solution based on unsupervised learning instead of a supervised one.

\section{A method for DRL assessment}
\label{sec:questions}

We introduce a method for gaining rapid and rough assessment of the data readiness levels of a given project. The method consists of a range of questions, intended to fuel the discussions between the stakeholders involved in a project with respect to its means and goals, as well as a simple way of visualizing the responses to the questions in order to bring attention to the areas that need more work. We expect to evolve the method in coming projects. So far, it has helped us to preemptively address some of the issues exemplified in Section \ref{sec:examples}; they are a good starting point in reaching the appropriate data readiness for solving real-world problems related to NLP.

\subsection{Pre-requisites}

The pre-requisites for applying the method are the following: there should be a clear business or research-related objective for the project to achieve; the objective lends itself to a data-driven solution; and, there is data available that presumably is relevant for the task.

The method should be scheduled for application at suitable points in time for the project, i.e., anytime the project enters a phase that relies on data and experimentation to make progress in the project plan. We suggest to apply the method at the very beginning of the project, as well as (at least) before entering the first round of empirical experiments with respect to the data at hand and the project's objective.


\subsection{Post-conditions}

The outcome of the method is two-fold: a visual representation of the Data Readiness Levels of the project at a specific point-in-time (as exemplified in Section \ref{sec:method-example}); and, the insight into the state of data achieved by venting the questions among the project's stakeholders.

\subsection{The questions}

The purpose of each question is to draw the stakeholders' attention to one aspect of the data readiness of the project. However, since not all questions are relevant to all types of projects some may be omitted depending on the characteristics of the project at hand.

Each of the fifteen questions below can be answered by one of four options: {\bf Don't know}, {\bf No}, {\bf Partially}, and {\bf Yes}, where {\bf Don't know} is always considered the worst possible answer, and {\bf Yes} as the answer to strive for. The admittedly very coarse grained answer scale is intended to serve as a guide in assessing the state of the project's data readiness, rather than as a definitive and elaborate tool for detailed assessment.

\subsubsection{Questions related to Band C}
Band C, that concerns the accessibility of data, is the band in that is the least dependent on the actual objective of the project, but clearing it is still required in order to make the project successful.

\begin{itemize}
    \item[Q1] {\bf Do you have programmatic access to the data?} The data should be made accessible to the people who are going to work with it, in a way that makes their work as easy as possible. This usually means programmatic access via an API, database, or spreadsheet.
    
    \item[Q2] {\bf Are your licenses in order?} In the case you plan on using data from a third-party provider, either commercial or via open access, ensure that the licences for the data permit the kind of usage that is needed for the current project. Furthermore, make sure you follow the Terms of Service set out by the provider.

    \item[Q3] {\bf Do you have lawful access to the data?} Make sure you involve the appropriate legal competence early on in your project. Matters regarding, e.g., personal identifiable information, and GDPR have to be handled correctly. Failing to do so may result in a project failure, even though all technical aspects of the project are perfectly sound.

    \item[Q4] {\bf Has there been an ethics assessment of the data?} In some use cases, such as when dealing with individuals' medical information, the objectives of the project require an ethics assessment. The rules for such a probe into the data are governed by strict rules, and you should consult appropriate legal advisors to make sure your project adheres to them.
    
    \item[Q5] {\bf Is the data converted to an appropriate format?} Apart from being accessible programmatically, and assessed with respect to licenses, laws, and ethics, the data should also be converted to a format appropriate for the potential technical solutions to the problem at hand. One particular challenge we have encountered numerous times, is that the data is on the format of PDF files. PDF is an excellent output format for rendering contents on screen or in print, but it is a terrible input format for data-driven automated processes (see, e.g., \cite{Panait2020} for examples).
    
\end{itemize}

\subsubsection{Questions related to Band B}
Band B concerns the validity of data. In pursuing projects with external parties, we have so far seen fairly few issues having to do with the validity of data. In essence, Band B is about trusting that the data format is what you expect it to be.

\begin{itemize}
    \item[Q6] {\bf Are the characteristics of the data known?} Are the typical traits and features of the data known? Perform an exploratory data analysis, and run it by all stakeholders in the project. Make sure to exemplify typical and extreme values in the data, and encourage the project participants to manually look into the data.
    
    \item[Q7] {\bf Is the data validated?} Ensure that the traits and features of the data make sense, and, e.g., records are deduplicated, noise is catered for, and that null values are taken care of.
    
\end{itemize}

\subsubsection{Questions related to Band A}

Band A concerns the utility of data. As such, it is tightly coupled to the objective of the project. In our experience, this is the most elusive data readiness level in that it requires attention every time the goal of a project changes.

\begin{itemize}
    \item[Q8] {\bf Do stakeholders agree on the objective of the current use case?} What problem are you trying to solve? The problem formulation should be intimately tied to a tangible business value or research hypothesis. When specifying the problem, make sure to focus on the actual need instead of a potentially interesting technology. The characteristics of the problem dictates the requirements on the data. Thus, the specification is crucial for understanding the requirements on the data in terms of, e.g., training data, and the need for manual labelling of evaluation or validation data. Only when you know the characteristics of the data, it will be possible to come up with a candidate technological approach to solve the problem.

    \item[Q9] {\bf Is the purpose of using the data clear to all stakeholders?} Ensure that all people involved in the project understands the role and importance of the data to be used. This is to solidify the efforts made by the people responsible for relevant data sources to produce data that is appropriate for the project's objective {\em and} the potential technical solution to address the objective.

    \item[Q10] {\bf Is the data sufficient for the current use case?} Given the insight into what data is available, consider the questions: What data is needed to solve the problem? Is that a subset of the data that is already available? If not: is there a way of getting all the data needed? If there is a discrepancy between the data available, and the data required to solve the problem, that discrepancy has to be mitigated. If it is not possible to align the data available with what is needed, then this is a cue to go back to the drawing board and either iterate on the problem specification, or collect suitable data. 

    \item[Q11] {\bf Are the steps required to evaluate a potential solution clear?} How do you know if you have succeeded? The type of data required to evaluate a solution is often tightly connected to the way the solution is implemented: if the solution is based on supervised machine learning, i.e., requiring labelled examples, then the evaluation of the solution will also require labelled data. If the solution depends on labelled training data, the process of annotation usually also results in the appropriate evaluation data. Any annotation effort should take into account the quality of the annotations, e.g., the inter-annotator agreement; temporal aspects of the data characteristics, e.g., information on when we need to obtain newly annotated data to mitigate model drift; and, the representativity of the data. \citet{tseng_best_2020} provide a comprehensive set of best-practices for managing annotation projects.

    \item[Q12] {\bf Is your organization prepared to handle more data like this beyond the scope of the project?} Even if the data processing in your organization is not perfect with respect to the requirements of machine learning, each project you pursue has the opportunity to articulate improvements to your organization's data storage processes. Ask yourself the questions: How does my organization store incoming data? Is that process a good fit for automatic processing of the data in the context of an NLP project, that is, is the data stored on a format that brings it beyond Band C ({\bf accessibility}) of the Data Readiness Levels? If not; what changes would need to be made to make the storage better?

    \item[Q13] {\bf Is the data secured?} Ensure that the data used in the project is secured in such a way that it is only accessible to the right people, and thus not accessible by unauthorized users. Depending on the sensitivity of the project, and thus the data, there might be a need to classify the data according to the security standards of your organization (e.g., ISO 27001), and implement the appropriate mechanisms to protect the data and project outcome. 
    
    \item[Q14] {\bf Is it safe to share the data with others?} In case the project aims to share its data with others, the risks of leaking sensitive data about, e.g., your organization's business plans or abilities have to be addressed prior to sharing it.
    
    \item[Q15] {\bf Are you allowed to share the data with others?} In case the project wishes to share its data, make sure you are allowed to do so according to the licenses, laws, and ethics previously addressed in the project.
\end{itemize}

\section{Example application of the method}
\label{sec:method-example}
For the purpose of exemplifying the use of the method described above, consider the fictitious case of project {\em Project}, an undertaking of a large organization with good experience of running conventional ICT projects, but little knowledge about data-driven NLP-based analysis. The actual subject matter and scope of {\em Project} is not important.

\begin{figure*}[!ht]
\begin{center}
\includegraphics[width=\linewidth]{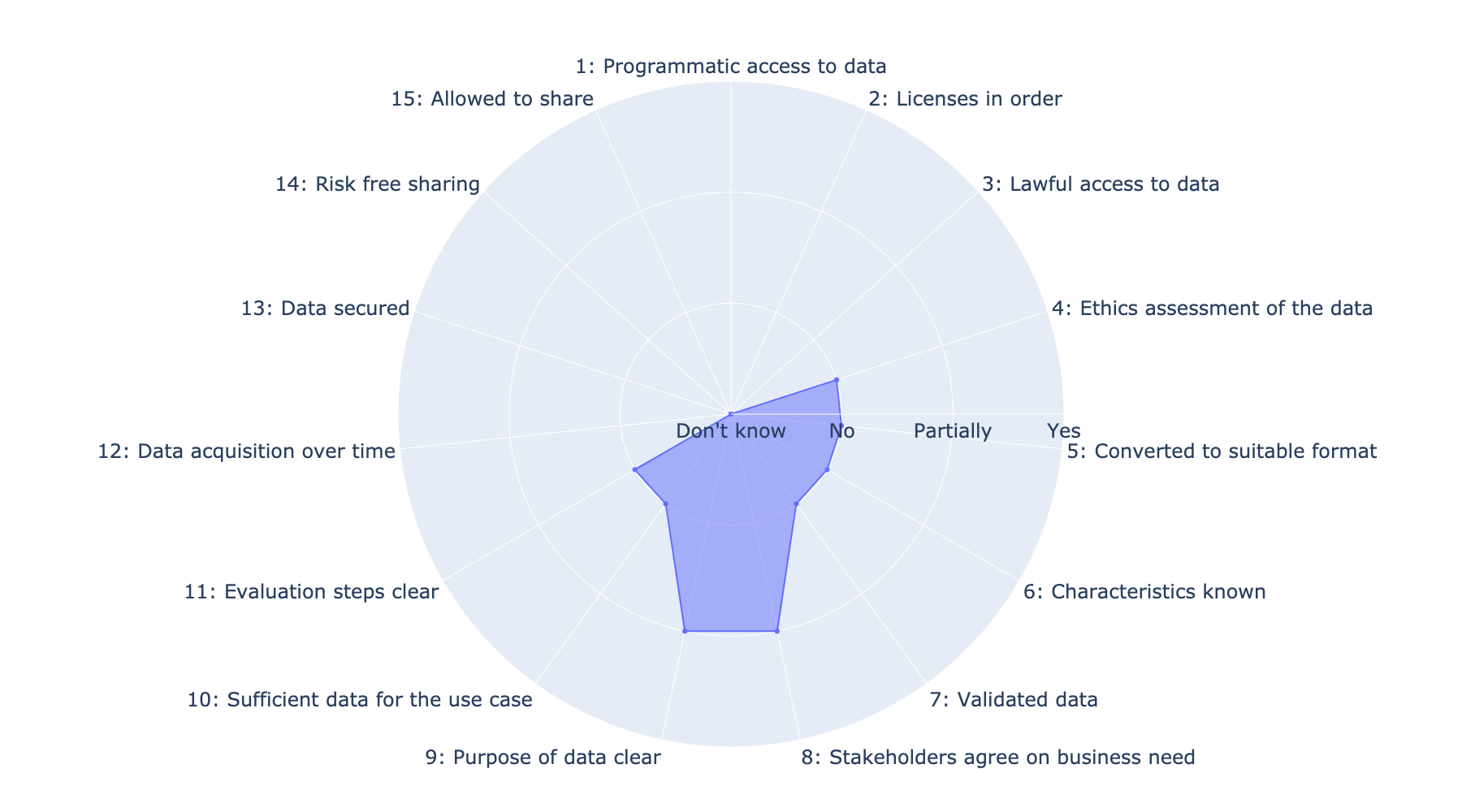}
\caption{The figure illustrates the state of data readiness at the beginning of the fictitious project.}
\label{fig:drl-assessment-ex-1}
\end{center}
\end{figure*}

\subsection{First application of the method}

When the project is initiated, the project manager involves its members in a session in which they recognize all fifteen questions as relevant for the project's objectives, then discusses each question, and agrees on the appropriate answer. When the session is over, the project manager plots the responses in a radar chart, as displayed in Figure \ref{fig:drl-assessment-ex-1}, in such a way that each of the questions answered is present in the chart, starting with Q1 ({\em Programmatic access to data}) up north, then progressing clock-wise with each question.\footnote{The code for generating radar charts as part of assessing the data readiness of your own project is available here: \url{https://github.com/fredriko/draviz}} The responses are plotted such that {\bf Don't know} (the worst answer) is located at the center of the chart, while {\bf Yes} (the best answer) is closest to the chart's edge. The aim of the assessment method is for the surface constituted by the enclosed responses to cover as large an area as possible. The reasons are simple; first, all stakeholders, in particular those in executive positions, will gain immediate visual insight into the state of the data for the project and, hopefully, feel the urge to act to increase the area; second, it is easy to visualize the project's progress at discrete points in time by overlaying two (or more) radar charts.\footnote{Overlaying radar charts with the purpose of comparing them only works when a small number of charts are involved; we are experimenting with parallel plots when the number of distinct charts exceeds 3.}

From Figure \ref{fig:drl-assessment-ex-1}, it can be seen that {\em Project}, at its incarnation, is not very mature with respect to data readiness. The area covered by the enclosed responses is small, and the number of unknowns are large. The only certainties resulting by the initial assessment of {\em Project}, are that: there has been no ethical assessment made of the data; the data has not been converted to a suitable format; no characteristics of the data are known; the data has not been validated; there is not sufficient data for the use case; and the way to evaluate the success of the project has yet to be defined. On the bright side, the stakeholder partially agrees on the objective of the project, and the purpose of the data.

\subsection{Second application of the method}

Fast forward to the second data readiness assessment of {\em Project}. In this case, it is scheduled to take place prior to the project embarking on the first rounds of empirical investigations of the state of data in relation to the project's business objective. The purpose of looking into the data readiness of the project at this stage, is to support the project manager in their work regarding prioritization, management of resources, and handling of expectations in relation to the project's progression and ability to reach its goals. 

\begin{figure*}[!ht]
\begin{center}
\includegraphics[width=\linewidth]{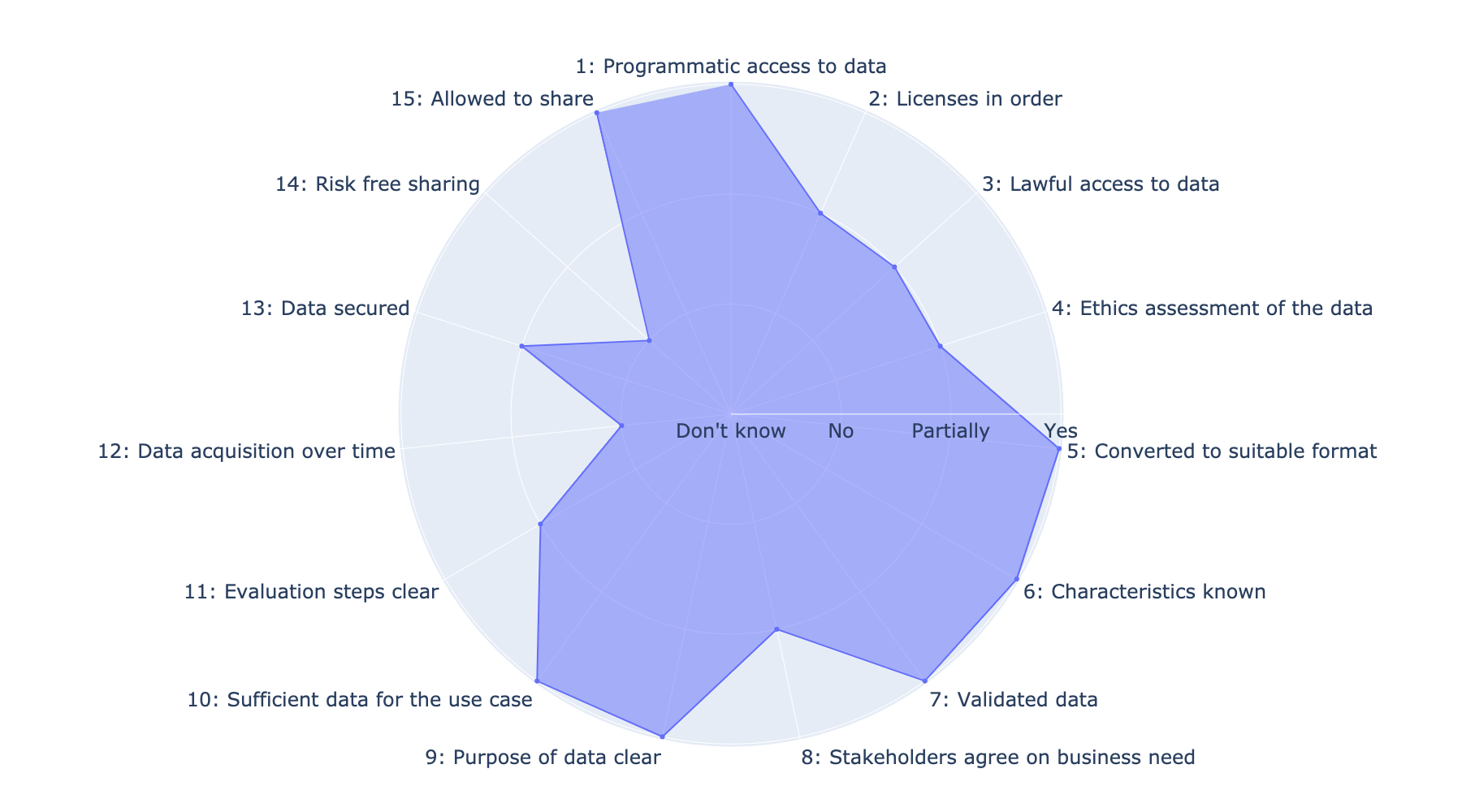}
\caption{The figure shows the corresponding state at the time where the project is ready to start making experiments based on the data.}
\label{fig:drl-assessment-ex-2}
\end{center}
\end{figure*}

Again, after all stakeholders have agreed on responses to the questions of the method, they are plotted in a radar chart. Figure \ref{fig:drl-assessment-ex-2} shows the state of the project after the second data readiness assessment. Progress has been made; the area covered by the responses is larger than it was at the initial assessment (Figure \ref{fig:drl-assessment-ex-1}). There are no unknowns left among the responses. Data is available and converted to a suitable format, its characteristics are known and the data format is generally trusted within the project. The fact that licenses, legal aspects of access, and ethics are not quite there yet does not, technically speaking, prohibit the project from moving on with the empirical investigation. However, these issues should be properly addressed before the project results are deployed to a general audience.

The stakeholders are still not in full agreement on the project's business objective, but they are aware of the purpose of the data, which has been deemed sufficient for the use case. Given the uncertainty with respect to the business objective, the steps required to evaluate proposed solutions are also unclear.

Beyond the scope of the project, the organization is not yet set up to in a way that is required to repeat and reproduce the findings of {\em Project} to future data, and data security is still work in progress. The project is allowed to share the data if it wishes to to so, but since management has decided to play it safe with respect to giving away too much information regarding the organization's future plans in doing so, it has been decided that data should not be share with external parties.



\section{Conclusions}
\label{sec:conclusions}

Research in NLP has never been more accessible; the impact of new results has the potential to reach far beyond the academic sphere. But with great power comes great responsibility. How can we 
foster a better uptake of research among public agencies, and in industry, and thereby gain valuable insight into the research directions that really matter? 
We introduce a method for assessing the Data Readiness Levels of a project, consisting of fifteen questions, and the accompanying means for visualizing the responses. We have utilized the proposed method and visualization technique in several projects with stakeholders in both the private and public sectors, and it has proven to be a very useful tool to improve the potential for successful application of NLP solutions to solve concrete business problems.

The method is a work-in-progress, and we thus invite researchers and practitioners in the NLP community to share their experience with respect to applied NLP research and data readiness at the following 
GitHub repository \url{https://github.com/fredriko/nlp-data-readiness}.

\bibliographystyle{acl_natbib}
\bibliography{main}

\begin{thebibliography}{24}
\expandafter\ifx\csname natexlab\endcsname\relax\def\natexlab#1{#1}\fi

\bibitem[{Afzal et~al.(2020)Afzal, C, Kesarwani, Mehta, and
  Patel}]{afzal_data_2020}
Shazia Afzal, Rajmohan C, Manish Kesarwani, Sameep Mehta, and Hima Patel. 2020.
\newblock \href {http://arxiv.org/abs/2010.07213} {Data {Readiness} {Report}}.
\newblock \emph{arXiv:2010.07213 [cs]}.
\newblock ArXiv: 2010.07213.

\bibitem[{Arnold et~al.(2019)Arnold, Bellamy, Hind, Houde, Mehta, Mojsilovic,
  Nair, Ramamurthy, Reimer, Olteanu, Piorkowski, Tsay, and
  Varshney}]{arnold_factsheets_2019}
Matthew Arnold, Rachel K.~E. Bellamy, Michael Hind, Stephanie Houde, Sameep
  Mehta, Aleksandra Mojsilovic, Ravi Nair, Karthikeyan~Natesan Ramamurthy,
  Darrell Reimer, Alexandra Olteanu, David Piorkowski, Jason Tsay, and Kush~R.
  Varshney. 2019.
\newblock \href {http://arxiv.org/abs/1808.07261} {{FactSheets}: {Increasing}
  {Trust} in {AI} {Services} through {Supplier}'s {Declarations} of
  {Conformity}}.
\newblock \emph{arXiv:1808.07261 [cs]}.
\newblock ArXiv: 1808.07261.

\bibitem[{Austin(2018)}]{austin_path_2018}
C.~C. Austin. 2018.
\newblock \href {https://doi.org/10.1109/BigData.2018.8622229} {A {Path} to
  {Big} {Data} {Readiness}}.
\newblock In \emph{2018 {IEEE} {International} {Conference} on {Big} {Data}
  ({Big} {Data})}, pages 4844--4853.

\bibitem[{Banke(2017)}]{NasaTrl2017}
Jim Banke. 2017.
\newblock {Technology Readiness Levels Demystified}.
\newblock URL:
  \url{https://www.nasa.gov/topics/aeronautics/features/trl_demystified.html}.
\newblock Accessed: 2021-01-11.

\bibitem[{Bender and Friedman(2018)}]{bender_data_2018}
Emily~M. Bender and Batya Friedman. 2018.
\newblock \href {https://doi.org/10.1162/tacl_a_00041} {Data {Statements} for
  {Natural} {Language} {Processing}: {Toward} {Mitigating} {System} {Bias} and
  {Enabling} {Better} {Science}}.
\newblock \emph{Transactions of the Association for Computational Linguistics},
  6:587--604.

\bibitem[{Ein-Dor et~al.(2020)Ein-Dor, Halfon, Gera, Shnarch, Dankin, Choshen,
  Danilevsky, Aharonov, Katz, and Slonim}]{ein-dor_active_2020}
Liat Ein-Dor, Alon Halfon, Ariel Gera, Eyal Shnarch, Lena Dankin, Leshem
  Choshen, Marina Danilevsky, Ranit Aharonov, Yoav Katz, and Noam Slonim. 2020.
\newblock \href {https://doi.org/10.18653/v1/2020.emnlp-main.638} {Active
  {Learning} for {BERT}: {An} {Empirical} {Study}}.
\newblock In \emph{Proceedings of the 2020 {Conference} on {Empirical}
  {Methods} in {Natural} {Language} {Processing} ({EMNLP})}, pages 7949--7962,
  Online. Association for Computational Linguistics.

\bibitem[{El-Kishky et~al.(2020)El-Kishky, Chaudhary, Guzmán, and
  Koehn}]{cc:ElKishkyChaudharyGuzmanKoehn:2020:ccaligned}
Ahmed El-Kishky, Vishrav Chaudhary, Francisco Guzmán, and Philipp Koehn. 2020.
\newblock \href {https://www.aclweb.org/anthology/2020.emnlp-main.480}
  {{CCA}ligned: {A} massive collection of cross-lingual web-document pairs}.
\newblock In \emph{Proceedings of the 2020 Conference on Empirical Methods in
  Natural Language Processing (EMNLP)}, pages 5960--5969, Online. Association
  for Computational Linguistics.

\bibitem[{Gebru et~al.(2020)Gebru, Morgenstern, Vecchione, Vaughan, Wallach,
  Daum{\'se}~III, and Crawford}]{gebru_datasheets_2020}
Timnit Gebru, Jamie Morgenstern, Briana Vecchione, Jennifer~Wortman Vaughan,
  Hanna Wallach, Hal Daum{\'se}~III, and Kate Crawford. 2020.
\newblock \href {http://arxiv.org/abs/1803.09010} {Datasheets for {Datasets}}.
\newblock \emph{arXiv:1803.09010 [cs]}.
\newblock ArXiv: 1803.09010.

\bibitem[{Harvey and Glocker(2019)}]{harvey2019standardised}
Hugh Harvey and Ben Glocker. 2019.
\newblock A standardised approach for preparing imaging data for machine
  learning tasks in radiology.
\newblock In \emph{Artificial Intelligence in Medical Imaging}, pages 61--72.
  Springer.

\bibitem[{Holland et~al.(2018)Holland, Hosny, Newman, Joseph, and
  Chmielinski}]{holland_dataset_2018}
Sarah Holland, Ahmed Hosny, Sarah Newman, Joshua Joseph, and Kasia Chmielinski.
  2018.
\newblock \href {http://arxiv.org/abs/1805.03677} {The {Dataset} {Nutrition}
  {Label}: {A} {Framework} {To} {Drive} {Higher} {Data} {Quality} {Standards}}.
\newblock \emph{arXiv:1805.03677 [cs]}.
\newblock ArXiv: 1805.03677.

\bibitem[{Lawrence(2017)}]{lawrence2017data}
Neil~D Lawrence. 2017.
\newblock Data readiness levels.
\newblock \emph{arXiv preprint arXiv:1705.02245}.

\bibitem[{Liang et~al.(2020)Liang, Zou, and Yu}]{liang_alice_2020}
Weixin Liang, James Zou, and Zhou Yu. 2020.
\newblock \href {https://doi.org/10.18653/v1/2020.emnlp-main.355} {{ALICE}:
  {Active} {Learning} with {Contrastive} {Natural} {Language} {Explanations}}.
\newblock In \emph{Proceedings of the 2020 {Conference} on {Empirical}
  {Methods} in {Natural} {Language} {Processing} ({EMNLP})}, pages 4380--4391,
  Online. Association for Computational Linguistics.

\bibitem[{Mackenzie et~al.(2020)Mackenzie, Benham, Petri, Trippas, Culpepper,
  and Moffat}]{cc:MackenzieBenhamPetriTrippasEtAl:2020:cc-news-en}
Joel Mackenzie, Rodger Benham, Matthias Petri, Johanne~R. Trippas, J.~Shane
  Culpepper, and Alistair Moffat. 2020.
\newblock \href {https://doi.org/10.1145/3340531.3412762} {{CC}-news-en: {A}
  large english news corpus}.
\newblock In \emph{Proceedings of the 29th ACM International Conference on
  Information \& Knowledge Management}, CIKM '20, pages 3077--3084, New York,
  NY, USA. Association for Computing Machinery.

\bibitem[{Paleyes et~al.(2020)Paleyes, Urma, and
  Lawrence}]{paleyes_challenges_2020}
Andrei Paleyes, Raoul-Gabriel Urma, and Neil~D. Lawrence. 2020.
\newblock \href {http://arxiv.org/abs/2011.09926} {Challenges in {Deploying}
  {Machine} {Learning}: a {Survey} of {Case} {Studies}}.
\newblock \emph{arXiv:2011.09926 [cs]}.
\newblock ArXiv: 2011.09926.

\bibitem[{Panait(2020)}]{Panait2020}
Bogdan Panait. 2020.
\newblock {What's so hard about PDF text extraction?}
\newblock URL: \url{https://filingdb.com/b/pdf-text-extraction}.
\newblock Accessed: 2021-05-11.

\bibitem[{Pelicon et~al.(2020)Pelicon, Pranjić, Miljković, Škrlj, and
  Pollak}]{pelicon_zero-shot_2020}
Andra{\v z} Pelicon, Marko Pranjić, Dragana Miljković, Blaž Škrlj, and
  Senja Pollak. 2020.
\newblock \href {https://doi.org/10.3390/app10175993} {Zero-{Shot} {Learning}
  for {Cross}-{Lingual} {News} {Sentiment} {Classification}}.
\newblock \emph{Applied Sciences}, 10(17):5993.
\newblock Number: 17 Publisher: Multidisciplinary Digital Publishing Institute.

\bibitem[{Polyzotis et~al.(2018)Polyzotis, Roy, Whang, and
  Zinkevich}]{polyzotis_data_2018}
Neoklis Polyzotis, Sudip Roy, Steven~Euijong Whang, and Martin Zinkevich. 2018.
\newblock Data {Lifecycle} {Challenges} in {Production} {Machine} {Learning}:
  {A} {Survey}.
\newblock \emph{SIGMOD Record}, 47(2):12.

\bibitem[{Settles(2012)}]{settles_active_2012}
Burr Settles. 2012.
\newblock \href
  {http://www.morganclaypoolpublishers.com/catalog_Orig/product_info.php?products_id=36}
  {\emph{Active {Learning}}}, volume 2012.
\newblock Morgan \& Claypool.

\bibitem[{Siddhant and Lipton(2018)}]{siddhant_deep_2018}
Aditya Siddhant and Zachary~C. Lipton. 2018.
\newblock \href {https://doi.org/10.18653/v1/D18-1318} {Deep {Bayesian}
  {Active} {Learning} for {Natural} {Language} {Processing}: {Results} of a
  {Large}-{Scale} {Empirical} {Study}}.
\newblock In \emph{Proceedings of the 2018 {Conference} on {Empirical}
  {Methods} in {Natural} {Language} {Processing}}, pages 2904--2909, Brussels,
  Belgium. Association for Computational Linguistics.

\bibitem[{Srivastava et~al.(2018)Srivastava, Labutov, and
  Mitchell}]{srivastava_zero-shot_2018}
Shashank Srivastava, Igor Labutov, and Tom Mitchell. 2018.
\newblock \href {https://doi.org/10.18653/v1/P18-1029} {Zero-shot {Learning} of
  {Classifiers} from {Natural} {Language} {Quantification}}.
\newblock In \emph{Proceedings of the 56th {Annual} {Meeting} of the
  {Association} for {Computational} {Linguistics} ({Volume} 1: {Long}
  {Papers})}, pages 306--316, Melbourne, Australia. Association for
  Computational Linguistics.

\bibitem[{Tseng et~al.(2020)Tseng, Stent, and Maida}]{tseng_best_2020}
Tina Tseng, Amanda Stent, and Domenic Maida. 2020.
\newblock \href {https://doi.org/10.13140/RG.2.2.34497.58727} {Best {Practices}
  for {Managing} {Data} {Annotation} {Projects}}.
\newblock \emph{arXiv:2009.11654 [cs]}.
\newblock ArXiv: 2009.11654.

\bibitem[{van Ooijen(2019)}]{van2019quality}
Peter~MA van Ooijen. 2019.
\newblock Quality and curation of medical images and data.
\newblock In \emph{Artificial Intelligence in Medical Imaging}, pages 247--255.
  Springer.

\bibitem[{Wenzek et~al.(2020)Wenzek, Lachaux, Conneau, Chaudhary, Guzmaán,
  Joulin, and Grave}]{wenzek_ccnet_2020}
Guillaume Wenzek, Marie-Anne Lachaux, Alexis Conneau, Vishrav Chaudhary,
  Francisco Guzmaán, Armand Joulin, and Edouard Grave. 2020.
\newblock \href {https://www.aclweb.org/anthology/2020.lrec-1.494} {{CCNet}:
  {Extracting} {High} {Quality} {Monolingual} {Datasets} from {Web} {Crawl}
  {Data}}.
\newblock In \emph{Proceedings of the 12th {Language} {Resources} and
  {Evaluation} {Conference}}, pages 4003--4012, Marseille, France. European
  Language Resources Association.

\bibitem[{Ye et~al.(2020)Ye, Geng, Chen, Chen, Xu, Zheng, Wang, Zhang, and
  Chen}]{ye_zero-shot_2020}
Zhiquan Ye, Yuxia Geng, Jiaoyan Chen, Jingmin Chen, Xiaoxiao Xu, SuHang Zheng,
  Feng Wang, Jun Zhang, and Huajun Chen. 2020.
\newblock \href {https://doi.org/10.18653/v1/2020.acl-main.272} {Zero-shot
  {Text} {Classification} via {Reinforced} {Self}-training}.
\newblock In \emph{Proceedings of the 58th {Annual} {Meeting} of the
  {Association} for {Computational} {Linguistics}}, pages 3014--3024, Online.
  Association for Computational Linguistics.

\end{thebibliography}

\end{document}